# Dither is Better than Dropout for Regularising Deep Neural Networks

Andrew J.R. Simpson [#1]

[#] Centre for Vision, Speech and Signal Processing, University of Surrey
Surrey, UK
[1] `Andrew.Simpson@surrey.ac.uk`

*Abstract*—**Regularisation of deep neural networks (DNN) during training is critical to performance. By far the most popular method is known as** *dropout*. **Here, cast through the prism of signal processing theory, we compare and contrast the regularisation effects of dropout with those of** *dither*. **We illustrate some serious inherent limitations of dropout and demonstrate that dither provides a more effective regulariser.**

## I. INTRODUCTION

In nonlinear signal processing, the use of additive noise prior to nonlinear processing (such as quantization or truncation) acts to decorrelate (or suppress) nonlinear distortion products. This process is known as *dithering* and can also be used in discrete signal processing to mitigate aliasing issues resulting from nonlinear distortion products which fall beyond the Nyquist limit.

Deep neural networks [1] may be interpreted as discrete (sampled) systems consisting of linear filters and nonlinear demodulation stages [2] and it has been suggested [3] that the inherent nonlinear distortion and aliasing contribute to problems of overfitting. Thus, in principle, if dither acts to suppress nonlinear distortion and aliasing it should also act to regularise a DNN.

At face value, *dropout* [4] appears somewhat compatible with dither and is known to be a useful regulariser. However, despite the cited motivation of 'preventing co-adaptation' [4], a coherent signal-processing-based rationale for dropout as regulariser has not emerged. In terms of sampling theory, although dropout acts similarly to dither in decorrelating nonlinear distortion products by perturbing the nonlinearity, it is not additive. A further critical difference between dropout and dither is that dropout discards a number of samples – a process that may be interpreted as stochastic decimation. One consequence of this is that dropout introduces nonlinear distortion and/or aliasing. In this paper, we illustrate that dither provides equivalent regularisation but with more rapid learning rates.

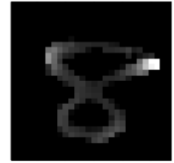

**Fig. 1. Example MNIST image.** We took the 28x28 pixel images and unpacked them into a vector of length 784 to form the input at the first layer of the DNN.

## II. METHOD

Regularisation is critical in the so-called 'small-data regime' – where the balance between parameters and data is skewed towards the parameters. For case study, we chose the well-known computer vision problem of hand-written digit classification using the MNIST dataset [5]. For the input layer we unpacked the images of 28x28 pixels into vectors of length 784. An example digit is given in Fig. 1. Pixel intensities were normalized to zero mean. Replicating Hinton's [6] architecture, but using the biased sigmoid activation function [2], we built a fully connected network of size 784x100x10 units, with the softmax output layer corresponding to the 10-way digit classification problem.

In order to place ourselves in the small-data regime, we used only the first 256 training examples of the MNIST dataset and tested on the full 10,000 test examples. We trained three versions of the model. The first version was trained without any regularisation. The second was trained with 50% dropout and the third version was trained with dither. For training with dither, uniform noise of unit scale and zero mean was added to the input (image only) data of each batch. The three classes of model were each independently instantiated and trained using SGD with batch sizes of 32, 64, 128 and 256 (i.e., 256 = full training set). Each separate model was trained for 100 full-sweep iterations of SGD (without momentum) and the test error computed (over the 10,000 test examples) at each iteration. For reliable comparison, each model was trained from the exact same random starting weights. A learning rate (SGD step size) of 1 was used for all training.

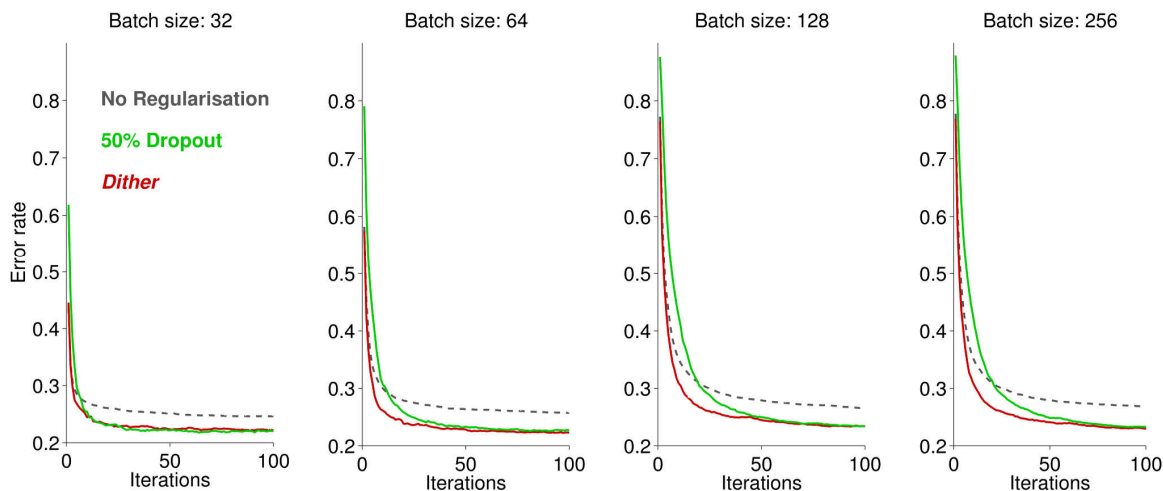

**Fig. 2. Regularisation during training: Dropout Versus Dither.** Test error function of SGD iterations, for the various batch sizes, for the un-regularised models (grey), the models trained with 50% dropout (green) and the models trained with dither (red).

## III. RESULTS

Fig. 2 plots the test-error rates, as a function of full-sweep SGD iterations, for the respective models of various batch sizes. Generally, performance is improved with regularisation and both dropout and dither converge to almost exactly the same point. However, dither learns faster and converges earlier. Performance is somewhat dependent upon batch size and is best for the batch size of 32. This tends to suggest that the data itself regularises best when averaged over batches of 32 and this probably relates to the nature of the data.

In summary, dither achieves the same ultimate degree of regularisation as dropout but the learning rate is superior to dropout. This most likely results from the additive nature of dither.

## IV. DISCUSSION AND CONCLUSION

In this paper, we have demonstrated that dither is a superior regulariser to dropout. We have rationalised the use of dither in terms of the regularisation provided taking the form of supression of nonlinear distortion and/or aliasing.

## ACKNOWLEDGMENT

AJRS did this work on the weekends and was supported by his wife and children.